%% file: main.tex
\documentclass[10pt,twocolumn,letterpaper]{article}

\usepackage{cvpr}              % To produce the CAMERA-READY version
\usepackage{bm}
\usepackage{multirow}


\definecolor{cvprblue}{rgb}{0.21,0.49,0.74}
\usepackage[pagebackref,breaklinks,colorlinks,allcolors=cvprblue]{hyperref}

\def\paperID{9787} % *** Enter the Paper ID here
\def\confName{CVPR}
\def\confYear{2025}

\title{GaussianUDF: Inferring Unsigned Distance Functions through 3D Gaussian Splatting}

\author{Shujuan Li\textsuperscript{\rm 1}, Yu-Shen Liu\textsuperscript{\rm 1}\thanks{Corresponding author: Yu-Shen Liu}, Zhizhong Han\textsuperscript{\rm 2}\\
\textsuperscript{\rm 1}School of Software, Tsinghua University, Beijing, China\\
\textsuperscript{\rm 2}Department of Computer Science, Wayne State University, Detroit, USA\\
{\tt\small lisj22@mails.tsinghua.edu.cn, liuyushen@tsinghua.edu.cn, h312h@wayne.edu}
}
\begin{document}
\maketitle

\begin{abstract}
Reconstructing open surfaces from multi-view images is vital in digitalizing complex objects in daily life. 
A widely used strategy is to learn unsigned distance functions (UDFs) by checking if their appearance conforms to the image observations through neural rendering. However, it is still hard to learn continuous and implicit UDF representations through 3D Gaussians splatting (3DGS) due to the discrete and explicit scene representation, i.e., 3D Gaussians. To resolve this issue, we propose a novel approach to bridge the gap between 3D Gaussians and UDFs. Our key idea is to overfit thin and flat 2D Gaussian planes on surfaces, and then, leverage the self-supervision and gradient-based inference to supervise unsigned distances in both near and far area to surfaces. To this end, we introduce novel constraints and strategies to constrain the learning of 2D Gaussians to pursue more stable optimization and more reliable self-supervision, addressing the challenges brought by complicated gradient field on or near the zero level set of UDFs. We report numerical and visual comparisons with the state-of-the-art on widely used benchmarks and real data to show our advantages in terms of accuracy, efficiency, completeness, and sharpness of reconstructed open surfaces with boundaries. Project page: \url{https://lisj575.github.io/GaussianUDF/}
\end{abstract}

\section{Introduction}
\label{sec:intro}
\begin{figure}
  \centering
  \includegraphics[width=1.0\linewidth]{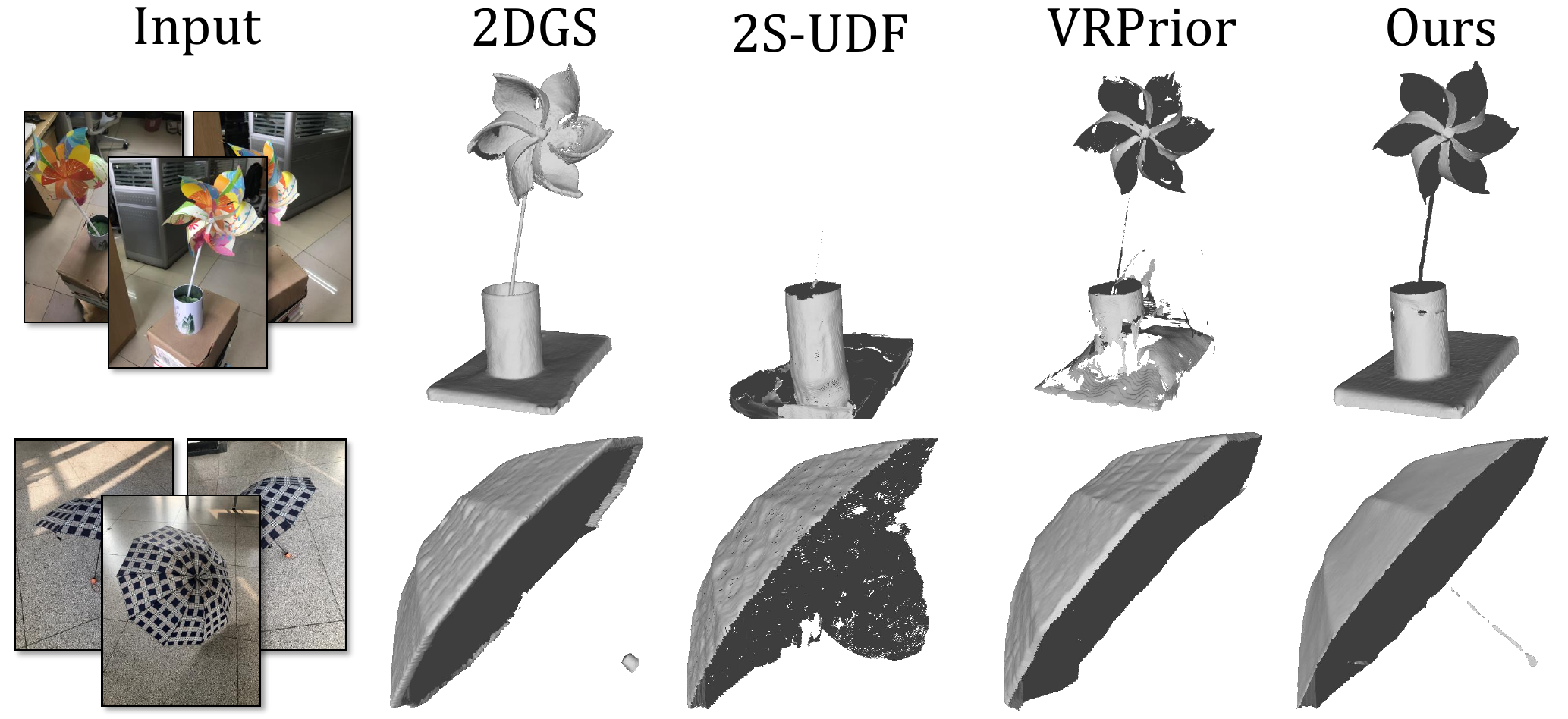}
   \caption{The comparisons with 2DGS~\cite{huang20242dgs}, 2S-UDF~\cite{deng20242sudf}, and VRPrior~\cite{zhang2024vrprior}. Our method recovers the most accurate open surfaces without artifacts.}
   \vspace{-10pt}
   \label{fig:first_compare}
   
\end{figure}

It is vital but still challenging to reconstruct shapes with open surfaces and sharp boundaries from multi-view images.
A widely used strategy is to learn implicit representations, such as unsigned distance functions (UDFs), by minimizing rendering errors of UDFs with respect to multi-view observations, such as RGB images~\cite{zhang2024vrprior, long2023neuraludf, deng20242sudf, liu2023neudf, meng2023neat}. This strategy shows promising results because of the advantages of both implicit representations and the neural volume rendering, i.e., the ability of reconstructing shapes with arbitrary topologies and the differentiability for back-propagating the gradients of rendering errors. 
Eventually, the open surfaces can be extracted from the zero level set of the learned UDF.

Recent methods~\cite{long2023neuraludf, liu2023neudf, deng20242sudf, meng2023neat, zhang2024vrprior} usually learn a UDF within a radiance field through the volume rendering introduced by NeRF~\cite{mildenhall2021nerf}. They infer unsigned distances at sampled points along rays emitted from views through an additional transformation which bridges the gap between the UDF and the radiance field. However, NeRF-based rendering is not efficient due to intersection finding in ray tracing. This makes 3D Gaussian Splatting (3DGS) a promising solution since rasterizing 3D Gaussians is not only differentiable but also faster than ray tracing in NeRF-based rendering. However, one obstacle coming from the discrete and explicit scene representations, i.e., 3D Gaussians, is that they are much different from the continuous and implicit radiance field. Therefore, how to overcome this obstacle is the most challenging problem to reveal complete, smooth, and continuous UDFs through 3DGS.

To resolve this problem, we introduce a novel approach to inferring UDFs from multi-view images through 3DGS, which can efficiently reconstruct high-quality surfaces with open structures as shown in Figure~\ref{fig:first_compare}. Our key idea is to constrain 3D Gaussians to represent surfaces directly, based on which we can estimate the unsigned distance field. Our novelty lies in two aspects: (1) the novel constraints that we imposed on the Gaussians, which overfits these Gaussians on surfaces, (2) and the ways of inferring unsigned distances with self-supervision and gradient-based inference. To this end, we use 2D Gaussians in the 3D space which are thin enough to approximate the surface. We also align these 2D Gaussians on the surface using the gradient field of the implicit function, which involves a UDF in the differentiable rasterization procedure. Meanwhile, we introduce self-supervision along the normal of 2D Gaussians to infer unsigned distances near the 2D Gaussians, and infer unsigned distances far away from the surface with the gradient field of the UDF. Our evaluations show that our method successfully bridges the gap between discrete Gaussians and continuous UDFs in a fully differentiable manner, leading to reconstructions of more accurate, complete, and continuous open surfaces than the state-of-the-art methods. Our contributions are summarized below.

\begin{itemize}
\item We present a novel approach to reconstruct open surfaces from multi-view images with 3DGS, which bridges the gap between continuous UDFs and discrete 3D Gaussians in a differentiable manner.
\item We introduce stable constraints to overfit 3D Gaussians on surfaces, and novel strategies to infer unsigned distances accurately in both near and far areas to the surface.
\item Our method produces state-of-the-art results in reconstructing shapes and scenes with open surfaces and sharp boundaries on the widely used benchmarks.
\end{itemize}

\input{sec/related_work}

\section{Method}

\begin{figure}
  \centering
  \includegraphics[width=1.0\linewidth]{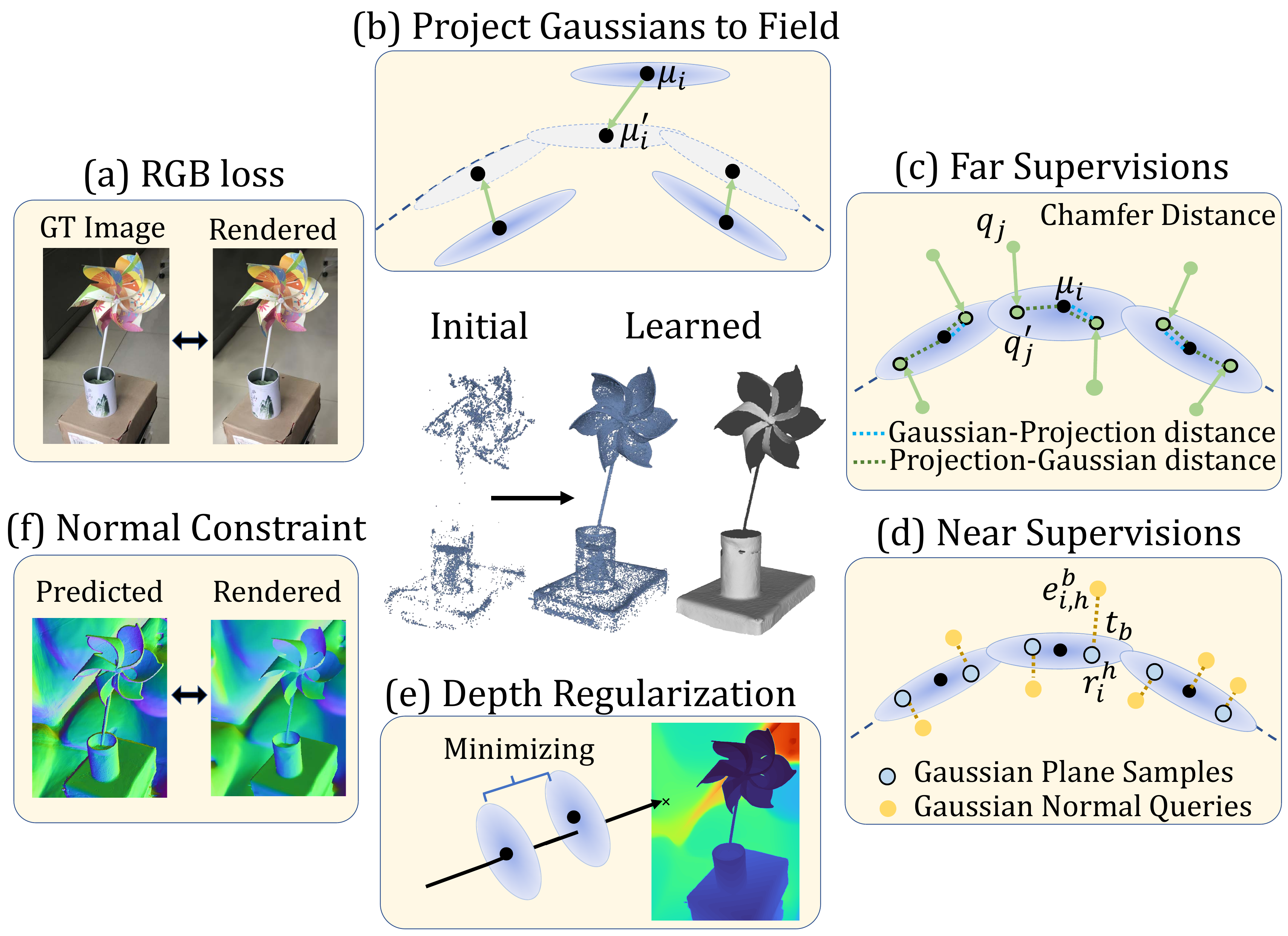}

  \caption{Overview of our method. (a) The UDF is optimized with the rendering process. To ensure that Gaussians can provide more accurate clues of the surfaces, (b) the Gaussians are projected to the zero level set of the UDF. (c) Projecting random queries to the Gaussian centers helps the UDF learn coarse distance fields which is far from surfaces. Moreover, (d) unsigned distances recovered near the Gaussian plane compensates for the sparsity of Gaussian centers. We adopt depth (e) and normal (f) regularization terms to make Gaussians align with surfaces well.}
   \label{fig:overview}
\end{figure}

\noindent\textbf{Overview. }Figure \ref{fig:overview} illustrates the framework of our approach. To overfit 3D Gaussians on surfaces, we follow 2DGS~\cite{huang20242dgs} to represent scenes using 2D Gaussians which are thin enough to represent open surfaces with sharp boundaries. We jointly infer a UDF $f$ and learn 2D Gaussians $\{\bm{g}_i\}^I_{i=1}$ by minimizing rendering errors with respect to the observations through splatting the $I$ 2D Gaussians $\{\bm{g}_i\}^I_{i=1}$. Besides the thin character of 2D Gaussians, we also leverage the gradient field of the UDF to align 2D Gaussians to the zero level set of the UDF, which ensures that these 2D Gaussians can represent the surface faithfully. Based on this representation, we set up self-supervision along the normal of 2D Gaussians to supervise the learning of UDF around the surface. Simultaneously, we also use the gradient field to infer unsigned distances, especially for the space far away from the surface. To this end, we also constrain the normal of 2D Gaussians and rendered depth images so that the 2D Gaussians can provide reliable self-supervisions and the gradient based inference for more accurate distance fields. 

\noindent\textbf{2D Gaussian Splatting. }We leverage the differentiable splatting introduced by 2DGS~\cite{huang20242dgs} to render 2D Gaussians into images. Each 2D Gaussian $\bm{g}_i$ has several learnable parameters including the center $\bm{\mu}_i\in\mathbb{R}^{1\times 3}$, the color $\bm{c}_i\in \mathbb{R}^{1\times 3}$, the opacity $\alpha_i$, the rotation matrix $\bm{r}_i\in \mathbb{R}^{3\times 3}$, and scaling factors $\bm{s}_i\in \mathbb{R}^{1\times 2}$, where $\bm{\mu}_i$ and $\bm{r}_i$ determine the location and pose of the Gaussian $\bm{g}_i$, $\bm{s}_i$ determines the variances along two axis of the Gaussian $\bm{g}_i$, the color $\bm{c}_i$ and the opacity $\alpha_i$ describe the appearance, and the last column of $\bm{r}_i$ represents the normal $\bm{n}_i$ of the flat Gaussian $\bm{g}_i$.

We render $\{\bm{g}_i\}$ into a RGB color $\bm{C}'(u,v)$ at each pixel $(u,v)$ on the rendered image $\bm{C}'$ using $\alpha$ blending through a differentiable splatting procedure,

\begin{equation}
\label{Eq:rendering1}
\bm{C}'(u,v)=\sum_{i=1}^I \bm{c}_i  \alpha_i p_i(u,v) \prod_{k=1}^{i-1} (1-\alpha_k p_k(u,v)),
\end{equation}
 where $p_i(u,v)$ is the probability of contributing to pixel $(u,v)$ from the projection of $\bm{g}_i$. Similarly, we can also render depth or normal maps by replacing the color with projection distances or the normal of 2D Gaussians in the above equation. We learn the Gaussians $\{\bm{g}_i\}$ by minimizing rendering errors with respect to the observations $\bm{C}$,
\begin{equation}
\label{Eq:rendering2}
L_{rgb}=||\bm{C}'(u,v)-\bm{C}(u,v)||_1.
\end{equation}
\noindent\textbf{Unsigned Distance Functions.} An unsigned distance function $f$ describes a distance field, indicating the distance $d$ to the nearest surface in a scene at an arbitrary location $\bm{q}=(x,y,z)$, i.e., $d=f(\bm{q})$. A gradient field can be derived from $f$, where the gradient $\nabla f(\bm{q})$ at each query $\bm{q}$ points to a direction that is far away from the nearest surface.

The gradient field of $f$ provides good clues to reveal surfaces which are indicated by the zero level set of $f$. Neural-Pull~\cite{ma2021npull} has shown that one can infer signed distances by pulling randomly sampled points along the gradient to the surface. However, a UDF produces a pretty complex gradient field near both sides of the surface, due to the absence of gradient on the surface. This fact becomes a serious problem in learning UDF from multi-view images.%, due to the explicit supervision indicating where surfaces are.

To resolve this issue, we employ two kinds of supervisions to infer unsigned distances with 2D Gaussians. One is to use the gradient field to pull queries onto the zero level set of $f$, which pays more attention to the space far away from surfaces. The other is to leverage the normal of the Gaussians to produce self-supervision covering the whole flat plane, which focuses on the area closed to surfaces.

\noindent\textbf{Self-supervision and Inference. }For the first supervision, we randomly sample $J$ queries $\{\bm{q}_j\}_{j=1}^J$ around the centers $\{\bm{\mu}_i\}$ of Gaussians $\{\bm{g}_i\}$ using the sampling strategy introduced in Neural-Pull~\cite{ma2021npull}. We project $\{\bm{q}_j\}$ onto the zero level set of $f$ below,

\begin{equation}
\label{Eq:pull}
\bm{q}_j'=\bm{q}_j-d_j \cdot \frac{\nabla f(\bm{q}_j)}{|\nabla f(\bm{q}_j)|},
\end{equation}
where $\bm{q}_j'$ is the projection of $\bm{q}_j$ and $d_j=f(\bm{q}_j)$ is the unsigned distance. We leverage the centers of Gaussians to supervise the projections,

\begin{equation}
\label{Eq:pullloss}
\begin{aligned}
L_{far}=&\frac{1}{J}\sum_{\bm{q}'\in\{\bm{q}_j'\}}\min_{\bm{\mu}\in\{\bm{\mu}_i\}}||\bm{q}'-\bm{\mu}||_2^2\\
&+\frac{1}{I}\sum_{\bm{\mu}\in\{\bm{\mu}_i\}}\min_{\bm{q}'\in\{\bm{q}_j'\}}||\bm{\mu}-\bm{q}'||_2^2,
\end{aligned}
\end{equation}

\noindent where $L_{far}$ evaluates the Chamfer distance between the set of projections $\{\bm{q}_j'\}$ and $\{\bm{\mu}_i\}$, encouraging the UDF $f$ to conform to the surface represented by the Gaussian centers. To relief the computational burden during optimization, we only use a batch of $\bm{g}_i$ and query points sampled around them to evaluate this loss in each iteration.

Gaussians are sparse in some regions, which limits their ability to represent surfaces, so relying solely on their centers with $L_{far}$ is inadequate.
Hence, the first supervision merely provide a coarse supervision which is helpful for inferring unsigned distances in areas far away from the surface. As a complement, our self-supervision will provide the second kind of supervision over the whole Gaussian plane near the surface.

\begin{figure}
  \centering
  \includegraphics[width=1.0\linewidth] {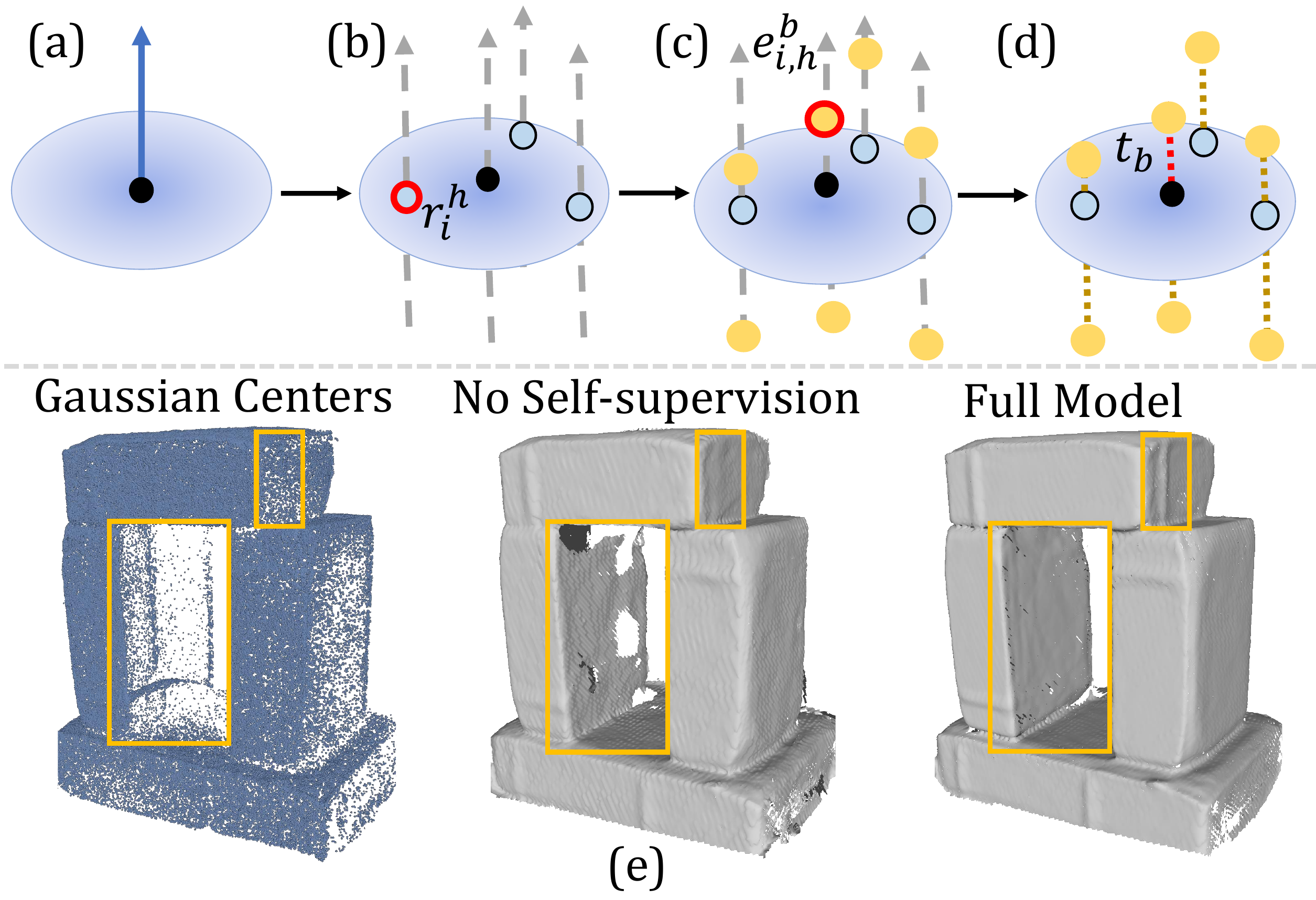}
   \caption{Self-supervision loss. For a Gaussian in (a),  (b) we first sample root point $r^h_i$ on the plane. (c) Then we randomly move the root point to position $e^b_{i,h}$ along or against the normal with a randomly sampled offset $t_b$. (d) We use $\{e^b_{i,h}, t_b\}$ as a training sample pair to train the UDF network. (e) The reconstructed meshes show that the 2D Gaussian planes provide more surface information for the UDF, which helps to fill the holes and capture more details.}
   \label{fig:self-supervision}
   \vspace{-5pt}
\end{figure}

Our self-supervision is illustrated in Figure~\ref{fig:self-supervision}. We set up the self-supervision using the normal $\bm{n}_i$ of each Gaussian $\bm{g}_i$ and the samples on its flat plane, which makes sure the Gaussian plane can cover enough space to overfit surfaces regardless of the sparsity of Gaussian centers. As shown in Figure~\ref{fig:self-supervision} (b), we sample root points $\{\bm{r}_i^h\}_{h=1}^H$ on the flat plane, and randomly sample samples $\{\bm{e}_{i,h}^b,t_b\}_{b=1}^B$ along the direction of normal $\bm{n}_i$ by $\bm{e}_{i,h}^b=\bm{r}_i^h+t_b \cdot \bm{n}_i/||\bm{n}_i||_2$, as shown in Fig.~\ref{fig:self-supervision} (c), where $t_b$ is randomly sampled from $[-T,T]$, which makes sure we have training samples on both sides of the Gaussian. We record $\bm{e}_{i,h}^b$ and $t_b$ as a training sample $\{\bm{e}_{i,h}^b,t_b\}$ in Fig.~\ref{fig:self-supervision} (d), where $t_b$ is regarded as the ground truth unsigned distances at $\bm{e}_{i,h}^b$. We will introduce another constraint $L_{norm}$ in Eq~\ref{Eq:normloss} to keep the normal of Gaussians orthogonal to surfaces, which also makes the self-supervision more reliable to use.

Eventually, we use $\{\bm{e}_i^b,t_b\}$ as self-supervision to train the UDF $f$ through a L1 loss,

\begin{equation}
\label{Eq:nearloss}
L_{near}=||f(\bm{e}_i^b)-t_b||_1,
\end{equation}

\begin{figure}
  \centering
  \includegraphics[width=1.0\linewidth] {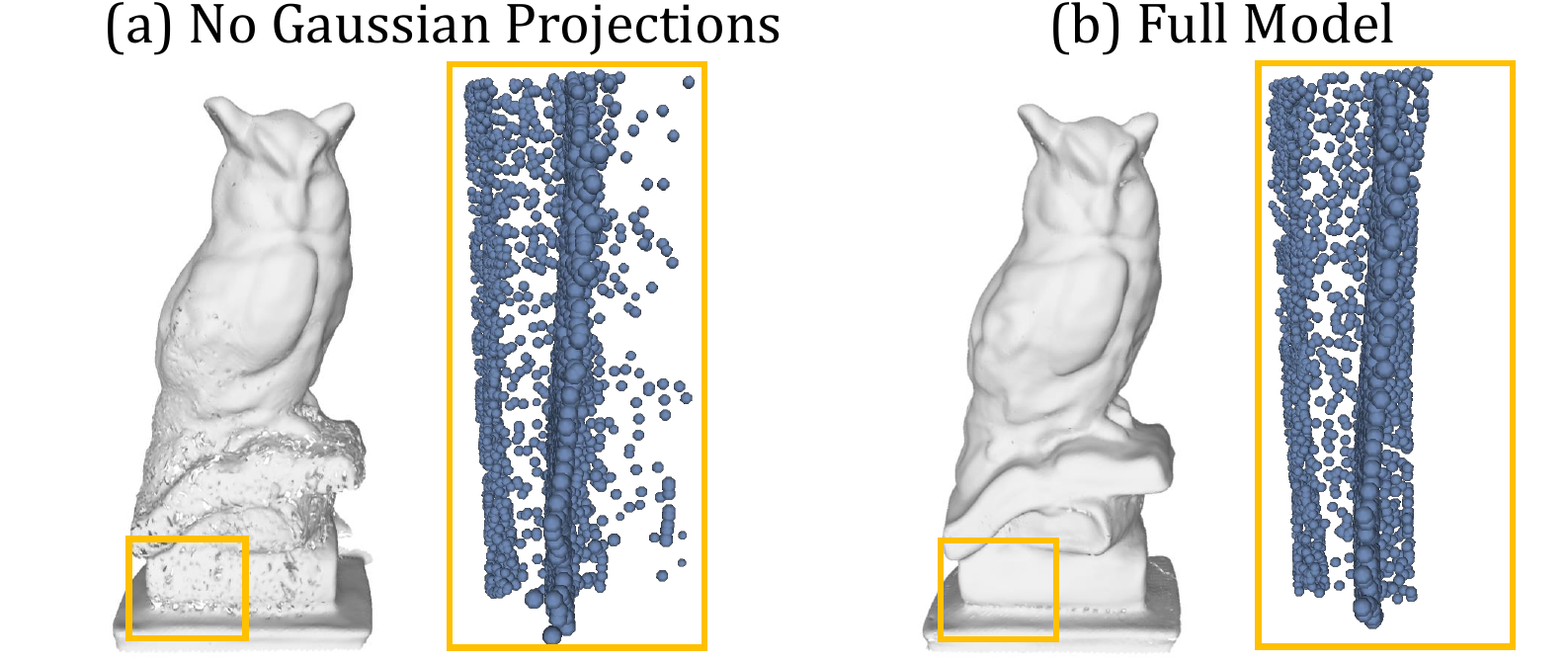}
   \caption{We project the Gaussian centers to the zero level set with a constraint, which makes the point cloud have less noises and the UDF have more accurate surface.}
   \vspace{-5pt}
   \label{fig:projection_loss}
\end{figure}

\begin{figure*}
  \centering
  \includegraphics[width=1.0\linewidth]{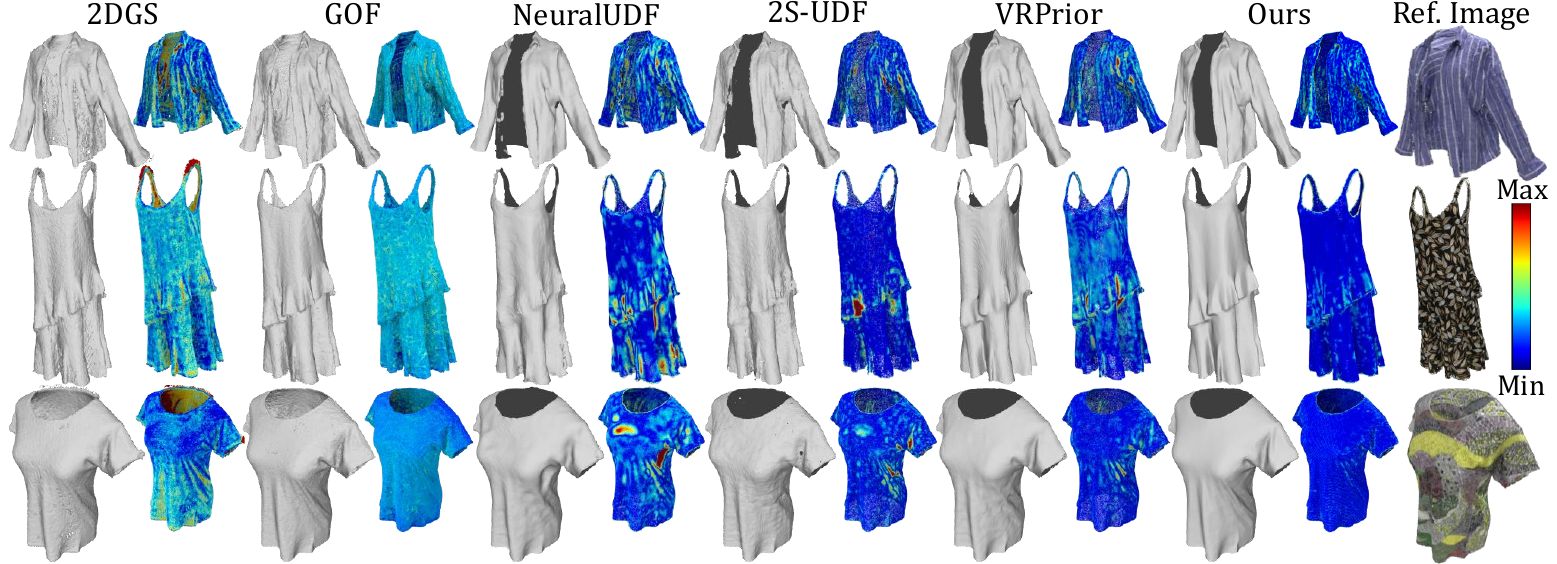}
    \caption{Qualitative comparison with 2DGS~\cite{huang20242dgs}, GOF~\cite{Yu2024GOF}, NeuralUDF~\cite{long2023neuraludf}, 2S-UDF~\cite{deng20242sudf}, and VRPrior~\cite{zhang2024vrprior} in DF3D~\cite{zhu2020df3d} dataset. Note that VRPrior needs additional depth images to learn priors. The dark color on meshes represents the back faces of open surfaces, and the error map is shown next to the mesh. Our method obtains more accurate surfaces and captures more details such as the folds in the clothing.}
   \label{fig:df3d_vis}
\end{figure*}

\begin{table*}
  \centering
  \small
  \begin{tabular}{@{}c|l|cccccccccccc|c|c@{}}
    \toprule
    &Method &30&92&117&133&164&204&300&320&448&522&591&598&Mean&Time \\
    \midrule
    \multirow{3}{*}{SDF}&NeuS\cite{wang2021neus}&3.18&4.82&4.78&4.99&3.73&5.71&5.89&2.21&5.89&3.60&2.44&5.13&4.36 & 5.7h \\
    &2DGS\cite{huang20242dgs}&3.79&3.66&4.24&3.75&3.91&4.01&4.02&3.74&3.51&3.89&3.21&4.01&3.81 & 6min\\
    &GOF\cite{Yu2024GOF}&3.15&2.47&2.49&2.23&2.38&2.65&2.40&2.41&2.14&3.00&2.18&2.37&2.49&47min\\
  %GSPull& \\
  \midrule
\multirow{4}{*}{UDF}&NeralUDF\cite{long2023neuraludf}&1.92&2.05&2.36&1.58&1.33&4.11&2.47&1.50&1.63&2.47&2.16&2.15&2.15&8.6h\\
 &2S-UDF \cite{deng20242sudf}&1.92&1.97&1.77&1.58&\textbf{1.32}&2.46&3.43&1.47&2.00&2.14&1.84&1.91&1.98&7.8h \\
 &VRPrior\cite{zhang2024vrprior}&\textbf{1.59}&1.73&2.06&1.63&1.44&2.07&1.66&1.60&\textbf{1.39}&2.14&\textbf{1.50}&1.67&1.71&9.2h\\
    &Ours&1.85&\textbf{1.69}&\textbf{1.18}&\textbf{1.32}&1.59&\textbf{1.59}&\textbf{1.51}&\textbf{1.27}&2.62&\textbf{1.65}&1.74&\textbf{1.22}&\textbf{1.60}&1.6h\\
    
    \bottomrule
  \end{tabular}
  \caption{Quantitative results of Chamfer Distance $(\times 10 ^{-3})$ of each object in DF3D~\cite{zhu2020df3d} dataset. Time is the average training time.}
  \vspace{-10pt}
  \label{tab:df3d_results}
\end{table*}

\noindent\textbf{Overfitting Gaussians to Surfaces. }Besides the thin character of 2D Gaussian, we also move 2D Gaussians to the zero level set of $f$, which ensures to overfit 2D Gaussians to surfaces. Since the gradient field nearby the zero level set of UDFs is very complicated, we do not directly pull the center $\bm{\mu}_i$ of 2D Gaussians $\bm{g}_i$ using Eq.~(\ref{Eq:pull}), which avoids the incorrect gradients that destablizes the optimization when most of 2D Gaussians are near the surface, as shown in Figure~\ref{fig:projection_loss} (a). We notice concurrent work~\cite{zhang2024gspull} that also involves gradients of SDF to constrain locations of Gaussians. But gradients of SDF near the zero level set are much more stable than UDF. Therefore, we propose to use an explicit constraint to project Gaussians on the zero level set of $f$. We run Eq.~(\ref{Eq:pull}) and stop back-propagating the gradient through $f$, obtaining the projection of Gaussian $\bm{\mu}_i'$. Then, we regard $\bm{\mu}_i'$ as target and minimize the distance to directly update the location $\bm{\mu}_i$ of Gaussians below, which stabilizes the optimization near the zero level set, as shown in Figure~\ref{fig:projection_loss} (b),

\begin{equation}
\label{Eq:locationloss}
L_{proj}=||\bm{\mu}_i'-\bm{\mu}_i||_2.
\end{equation}

\noindent\textbf{Constraints on Depth and Normals. }To make all 2D Gaussians get closer to the surface, we also adopt a depth distortion loss~\cite{huang20242dgs} to constrain Gaussian positions. Along each ray, we monitor the depth of intersections to Gaussians, and constrain their interval between two intersections,

\begin{equation}
\label{Eq:depthloss}
L_{depth}=\sum_{k1,k2} g_{k1}g_{k2}|z_{k1}-z_{k2}|,
\end{equation}

\noindent where $g_{k1}=\alpha_{k1}p_{k1}(u,v)\prod_{k=1}^{k1-1} (1-\alpha_kp_k(u,v))$. 

Furthermore, to make the self-supervision more reliable, we add supervision on the normals of Gaussians $\bm{n}_i$ like ~\cite{huang20242dgs}. We estimate normal maps from the depth gradients on the rendered depth images. Along each ray, we align the normal $\bm{n}_i$ of Gaussians hit by the ray with the estimated normal $\bm{N}_i$ on the rendered depth maps,

\begin{equation}
\label{Eq:normloss}
L_{norm}=\sum_{k} g_{k}(1-\bm{n}_k^T\bm{N}_k).
\end{equation}

\noindent\textbf{Loss Function. }We optimize 2D Gaussians in a scene by minimizing the following loss function,

\begin{equation}
\label{Eq:loss}
\begin{aligned}
L=&(1-\lambda_1)L_{rgb}+\lambda_1 L_{ssim}+\lambda_2 L_{far}+\lambda_3 L_{near} \\
&+\lambda_4 L_{proj}+\lambda_5 L_{depth}+\lambda_6 L_{norm},
\end{aligned}
\end{equation}

\noindent where $L_{ssim}$ is a rendering quality loss from 3DGS~\cite{kerbl20233dgs}, and all loss terms are balanced by weights $\lambda_{1-6}$.

\begin{figure*}
  \centering
  \includegraphics[width=1.0\linewidth]{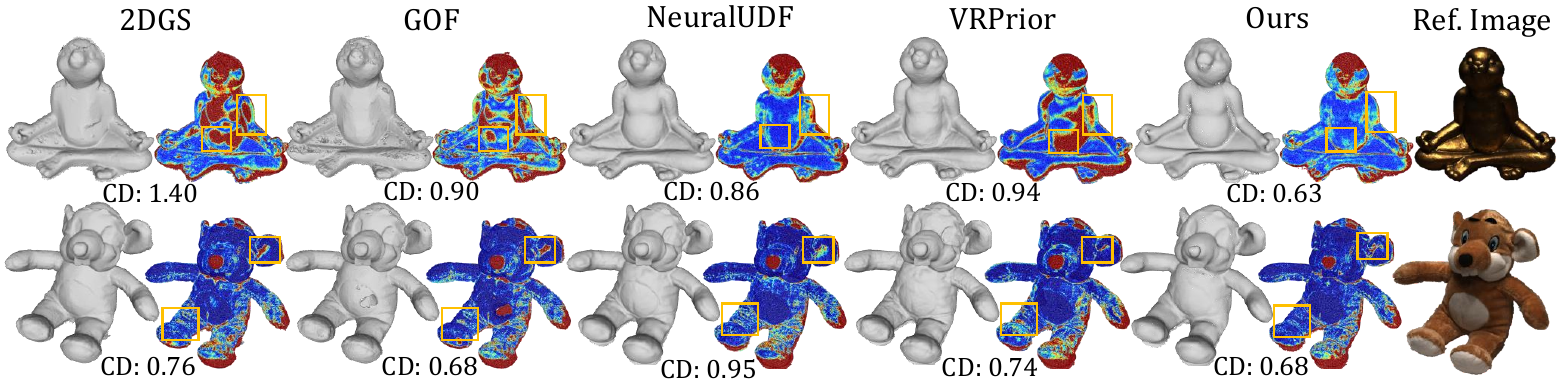}
    \vspace{-15pt}
   \caption{Visual comparisons of reconstruction and error maps in reconstruction on DTU \cite{jensen2014dtu} dataset. Larger errors are shown in warmer colors. Our method obtains visual-appealing results with small errors.}
   \vspace{-10pt}
   \label{fig:dtu_vis}
\end{figure*}

\input{sec/experiment}

\section{Conclusion}
We introduce an approach to reconstructing shapes with open surfaces and sharp boundaries from multi-view images with 3DGS. Our method can not only benefit from the high training efficiency of 3DGS, but also recover more accurate, complete, and continuous UDFs from discrete 3D Gaussians. The proposed constraints effectively overfit 3D Gaussians on surfaces, based on which our strategies for unsigned distance inference can recover high fidelity unsigned distance fields. Our evaluations justify the effectiveness of each module, and show advantages over the latest methods in terms of accuracy, completeness, and sharpness on reconstruction with open surfaces. 
\section{Acknowledgement}
We sincerely thank Wenyuan Zhang for his valuable suggestions. 
This work was partially supported by Deep Earth Probe and Mineral Resources Exploration -- National Science and Technology Major Project (2024ZD1003405), and the National Natural Science Foundation of China (62272263).

{
    \small
    \bibliographystyle{ieeenat_fullname}
    \bibliography{main}
}

\end{document}

%\input{sec/X_suppl}
%\maketitle

\maketitlesupplementary

%\title{Supplementary Material}

\section{Source Codes}
We provide our demonstration code as a part of our supplementary materials. We will release the source code, data and pretrained models upon acceptance.

\section{Implementation Details}
\textbf{Details for Self-supervision. }For sampling root points, we first sample random points from a standard normal distribution and then transform them onto the 2D Gaussian's plane using the corresponding rotation matrix, variance, and mean. The offset direction of these root points is determined by the normal vector of the Gaussian that generated them. For Gaussians whose maximum scale exceeds a certain threshold, we sample more root points. Specifically, we set this threshold to three times the mean of the maximum scales of all Gaussians.

\noindent\textbf{Data Preparation. }To validate the effectiveness of our method on real-world data, we additionally captured four scenes with open surfaces. We took 50–90 images per scene with a smartphone, each image with a resolution of $4032\times3024$. Following the processing procedure of NeuS~\cite{wang2021neus}, we first used COLMAP~\cite{schonberger2016structure} to estimate a sparse point cloud, then manually adjusted the regions of interest, and finally estimated the corresponding camera parameters for training.

\begin{figure}[t]
  \centering
  \includegraphics[width=1.0\linewidth]{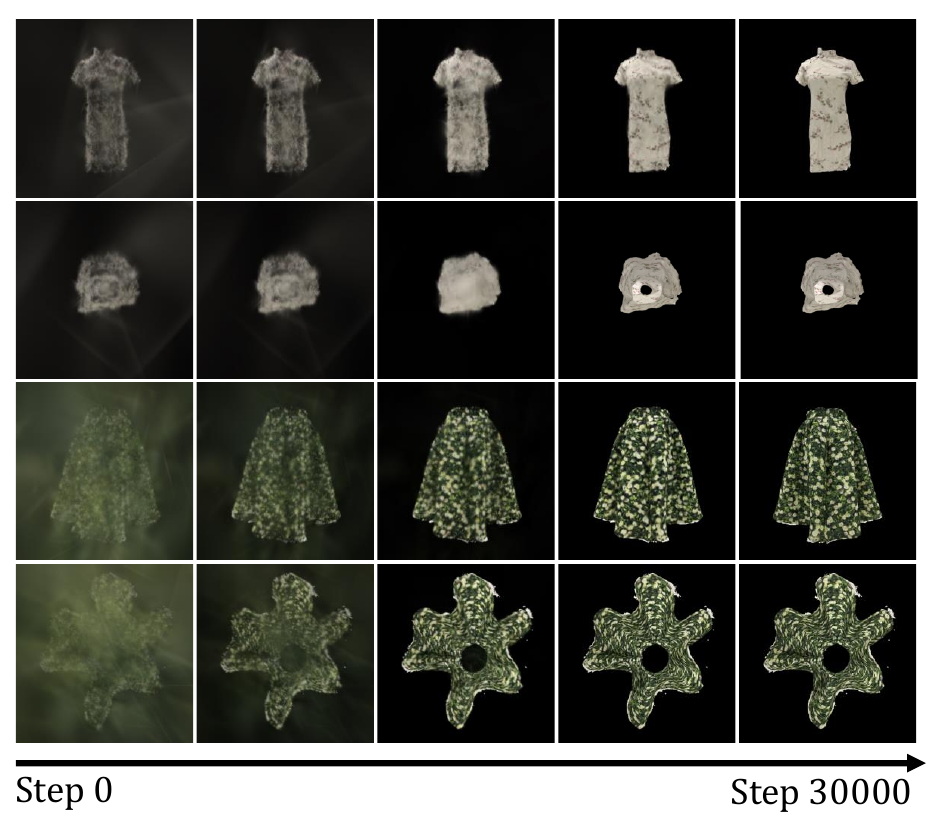}
   \caption{Training process of rendering image.}
   \label{fig:render_process}
\end{figure}

\section{Visualization for Optimization Process}
\begin{figure}[t]
  \centering
  \includegraphics[width=1.0\linewidth]{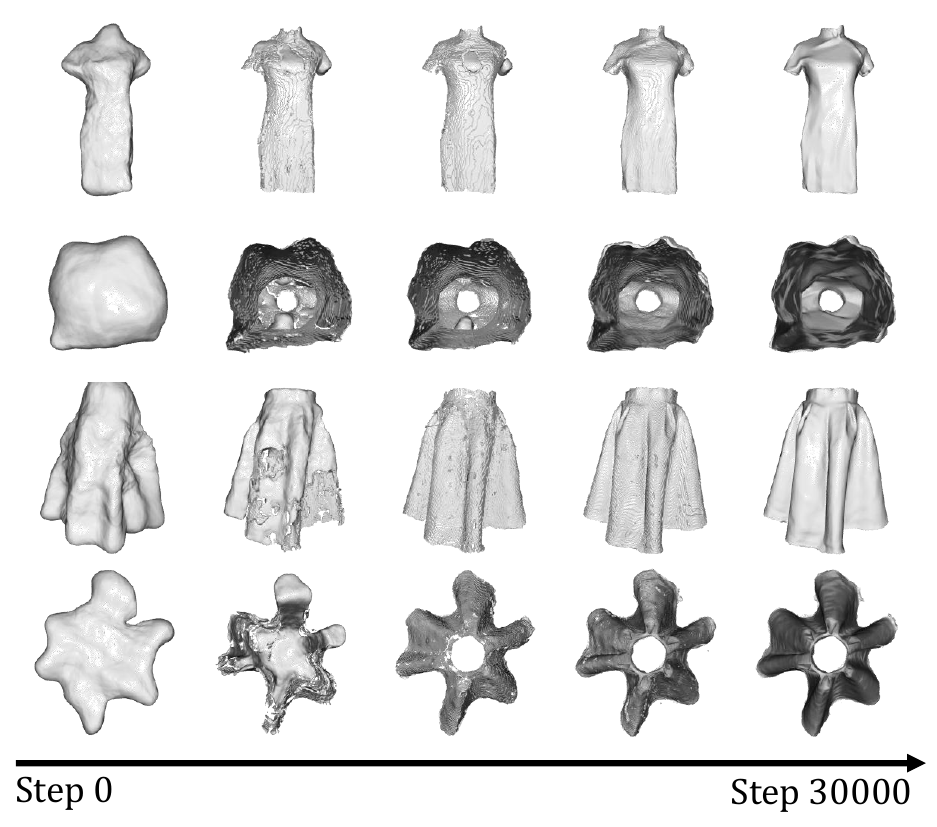}
   \caption{Training process of reconstructing meshes.}
   \label{fig:recon_process}
\end{figure}

\begin{figure}
  \centering
  \includegraphics[width=1.0\linewidth]{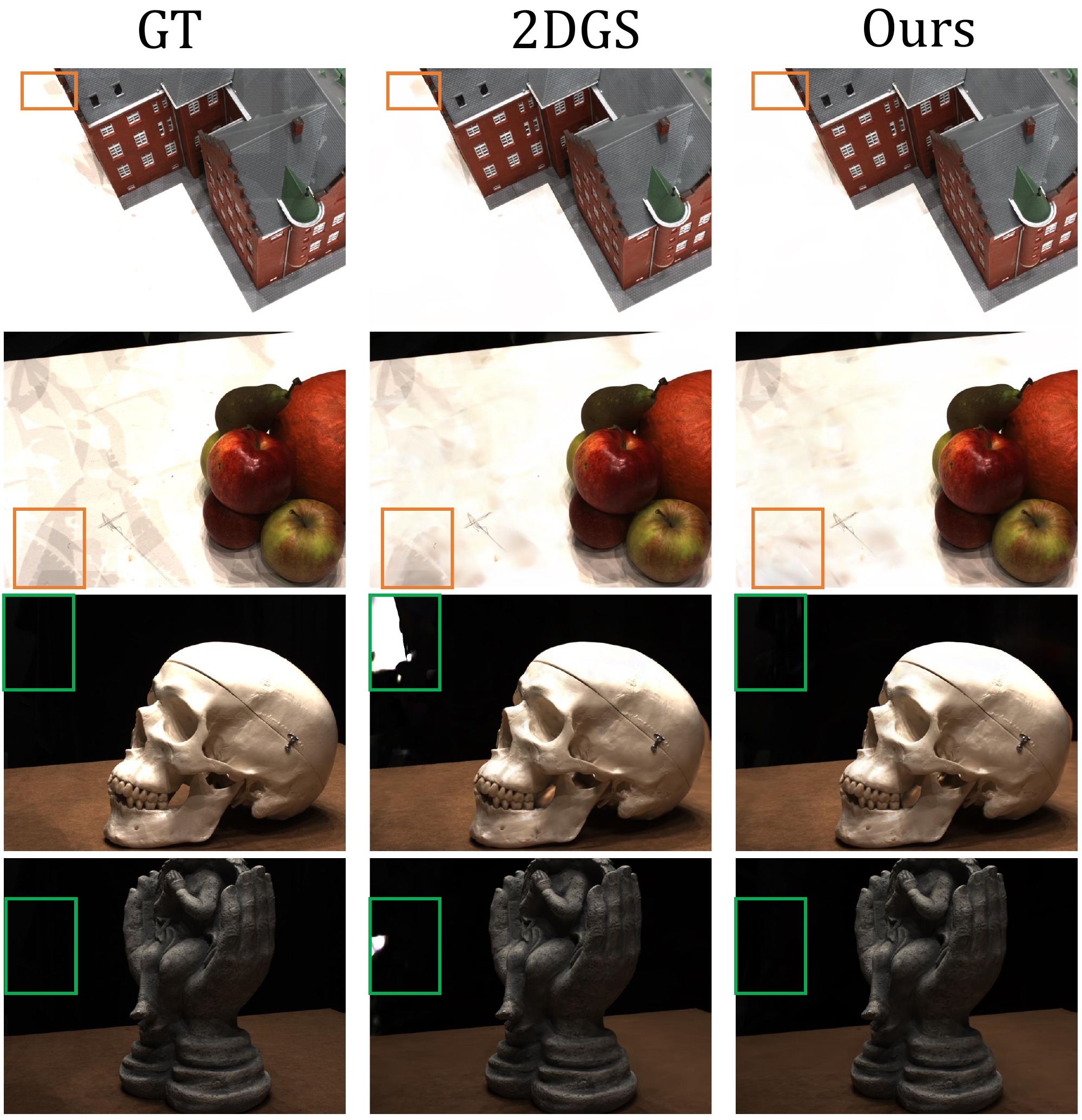}
   \caption{Rendering quality analysis on DTU~\cite{jensen2014dtu} dataset. Our method renders images with fewer floating artifacts.}
   \label{fig:dtu_render_compare}
\end{figure}

\begin{figure*}
  \centering
  \includegraphics[width=1.0\linewidth]{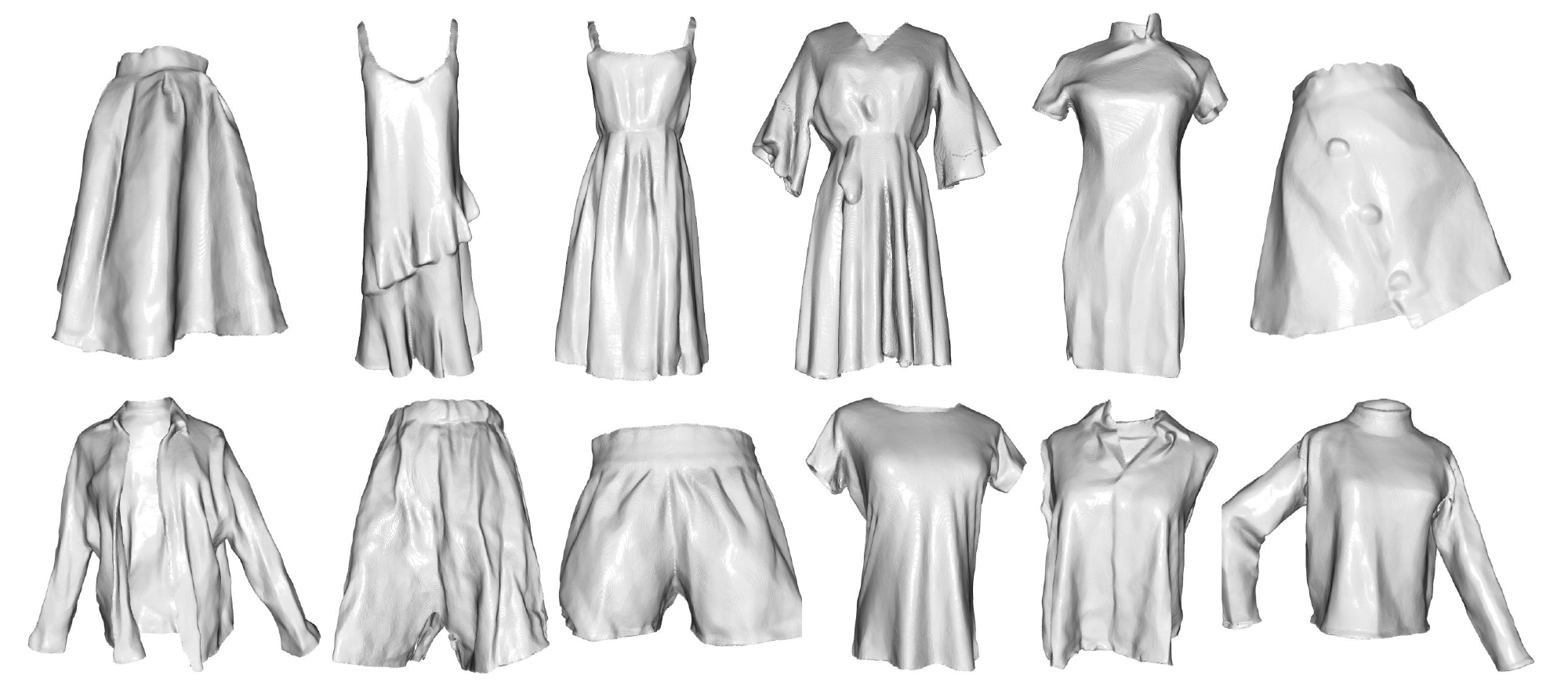}

   \caption{All reconstruction results on DF3F~\cite{zhu2020df3d} dataset. Our method accurately reconstructs the open surfaces.}
   \label{fig:df3d_more}
\end{figure*}

\begin{figure*}
  \centering
  \includegraphics[width=1.0\linewidth]{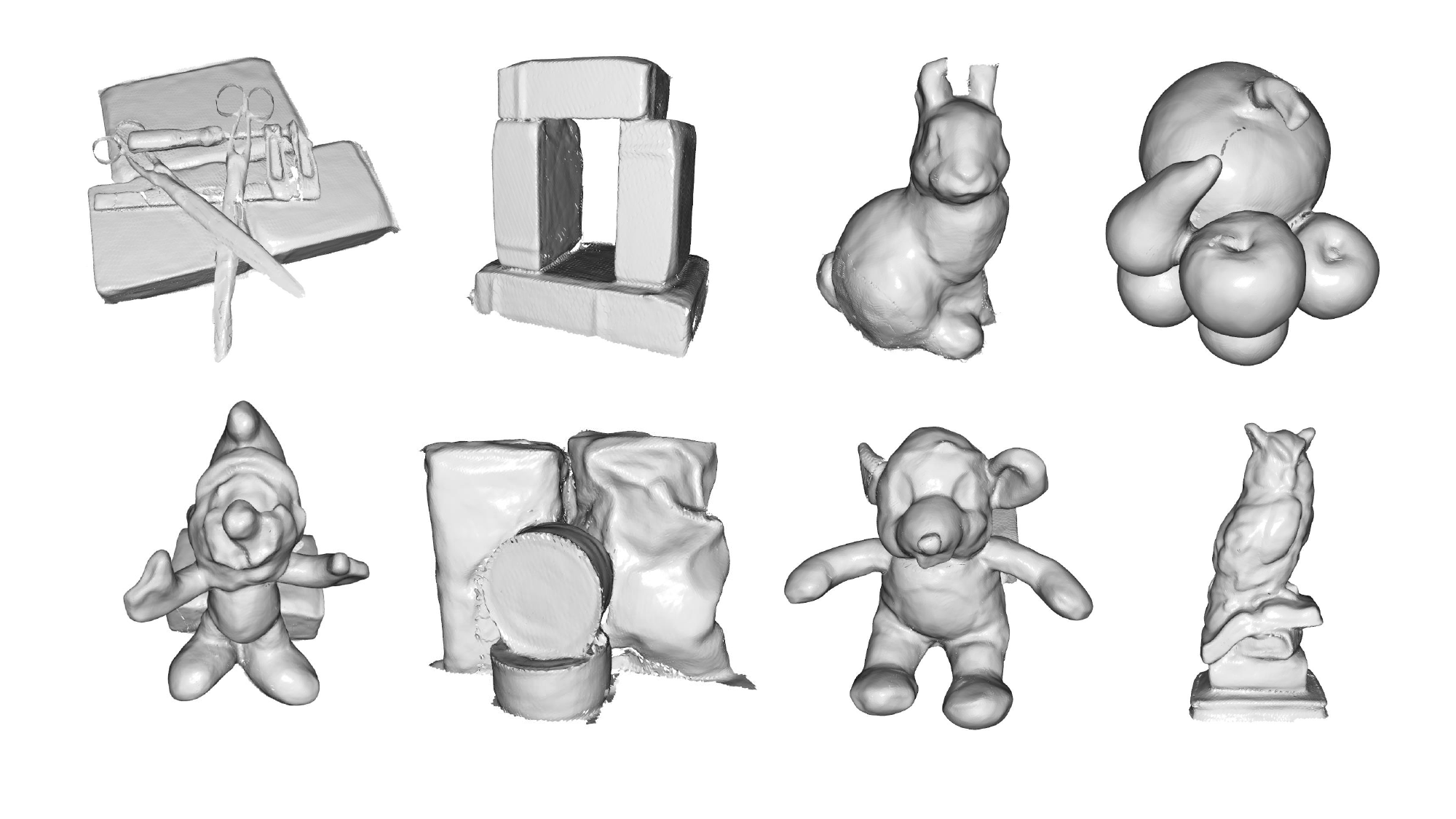}

   \caption{More reconstruction results on DTU~\cite{jensen2014dtu} dataset.}
   \label{fig:dtu_more}
\end{figure*}

\begin{table*}
  \centering
  \small
  \begin{tabular}{@{}l|ccccccccccccccc|c@{}}
    \toprule
    Method & 24 & 37 & 40 & 55 & 63 & 65 & 69 & 83 & 97 & 105 & 106 & 110 & 114& 118& 122 & Mean \\
    \midrule
    NeuS\cite{wang2021neus}&1.00&1.37&0.93&0.43&1.10&\textbf{0.65}&\textbf{0.57}&1.48&1.09&0.83&\textbf{0.52}&1.20&\underline{0.35}&\textbf{0.49}&0.54&0.84 \\
    3DGS\cite{kerbl20233dgs}&2.14&1.53&2.08&1.68&3.49&2.21&1.43&2.07&2.22&1.75&1.79&2.55&1.53&1.52&1.50&1.96 \\
    SuGaR\cite{guedon2024sugar}&1.47&1.33&1.13&0.61&2.25&1.71&1.15&1.63&1.62&1.07&0.79&2.45&0.98&0.88&0.79&1.33\\
    2DGS\cite{huang20242dgs}&\textbf{0.48}&0.91&\underline{0.39}&0.69&1.01&0.83&0.81&1.36&1.27&0.76&0.70&1.40&0.40&0.76&0.52&0.80 \\
  
  GSPull\cite{zhang2024gspull}&0.51&\textbf{0.56}&0.46&\underline{0.39}&\textbf{0.82}&0.67&0.85&1.37&1.25&0.73&\underline{0.54}&1.39&\underline{0.35}&0.88&\textbf{0.42}&0.75 \\
  GOF\cite{Yu2024GOF}&\underline{0.50}&0.82&\textbf{0.37}&\textbf{0.37}&1.12&0.74&0.73&\textbf{1.18}&1.29&\textbf{0.68}&0.77&0.90&0.42&0.66&\underline{0.49}&0.74\\
  \midrule
 NeuralUDF&0.69&1.18&0.67&0.44&0.90&\underline{0.66}&0.67&1.32&\textbf{0.94}&0.95&0.57&\underline{0.86}&0.37&0.56&0.55&0.75\\
 2S-UDF*\cite{deng20242sudf} &--  &0.89&--  &0.55&--  &0.68&0.88&--  &1.15&\underline{0.70}&0.74&--  &0.41&0.61&0.51&0.71\\
 VRPrior\cite{zhang2024vrprior}&0.55&0.95&0.45&0.43&0.94&0.75&\underline{0.61}&1.40&\underline{1.01}&0.74&0.60&0.94&\textbf{0.31}&\underline{0.50}&0.50&\underline{0.71}\\
 
    Ours&0.62&\underline{0.67}&0.43&0.42&\underline{0.83}&0.86&0.72&\underline{1.20}&1.03&\textbf{0.68}&0.61&\textbf{0.63}&0.43&0.56&0.52&\textbf{0.68}\\

    \bottomrule
  \end{tabular}
  \caption{Numerical comparisons in terms of CD on DTU~\cite{jensen2014dtu} dataset. Bold and underlined numbers indicate the first and second best performance, respectively. *We use the metrics reported in the paper of 2S-UDF.}
  \label{tab:dtu_results_detail}
\end{table*}

\begin{table*}
  \centering
  \small
  \begin{tabular}{@{}l|ccccccccccccccc|c@{}}
    \toprule
   Setting & 24 & 37 & 40 & 55 & 63 & 65 & 69 & 83 & 97 & 105 & 106 & 110 & 114& 118& 122 & Mean \\
    \midrule
    Only Far&0.76&0.95&0.68&0.73&0.92&1.14&1.09&1.4&1.35&0.81&0.99&1.41&0.96&0.84&0.77&0.99\\
    Far$\And$Near&0.6&0.69&0.4&0.48&0.84&0.84&0.84&1.37&1.13&0.7&0.76&1.16&0.49&0.74&0.62&0.78\\
    Far$\And$Proj&0.7&0.73&0.6&0.54&0.85&1.3&0.99&1.25&1.2&0.7&0.96&1.29&0.57&0.81&0.64&0.88\\
    w/o Warp&0.65&0.69&0.41&0.48&0.82&0.85&0.76&1.21&1.08&0.67&0.65&1.14&0.51&0.65&0.53&0.74 \\
  w/o Near&0.72&0.79&0.53&0.4&0.8&1.02&0.91&1.29&1.02&0.72&0.68&0.79&0.65&0.67&0.6&0.77\\
  w/o Proj&0.81&0.85&0.49&0.5&0.87&0.87&0.79&1.25&1.1&0.89&0.58&0.67&0.49&0.6&0.64&0.76\\
  \midrule
 Full Model&0.62&0.67&0.43&0.42&0.83&0.86&0.72&1.2&1.03&0.68&0.61&0.63&0.43&0.56&0.62&0.68\\

    \bottomrule
  \end{tabular}
  \caption{The detailed CD on DTU~\cite{jensen2014dtu} dataset for ablation studies.}
  \label{tab:ablation_results_detail}
\end{table*}

We show the training process in Figure~\ref{fig:render_process} and Figure~\ref{fig:recon_process}. During training, we first optimize the rendering loss to learn the Gaussian representations, then use the far supervision to guide the implicit field in learning a coarse shape from the Gaussian centers. Based on this, we incorporate the projection loss to optimize the Gaussian point cloud, leading to thinner surfaces and much less noises around geometry structures. Meanwhile, we add the self-supervision to achieve more accurate and complete reconstruction. Please refer to the whole process in our video for more details.

\section{Novel View Synthesis}

\begin{table}
    \centering
    \begin{tabular}{c|cc|cc}
        \toprule
            & \multicolumn{2}{c|}{Train} & \multicolumn{2}{c}{Test} \\
         Method & 2DGS & Ours & 2DGS & Ours\\
         \midrule
         SSIM  $\uparrow$ & 0.938&0.937&0.903&0.904 \\
         PSNR $\uparrow$ & 34.48&33.78&28.37&28.40\\
         LPIPS $\downarrow$ & 0.169&0.171&0.195&0.195\\
         \bottomrule
    \end{tabular}
    \caption{Rendering quality on DTU~\cite{jensen2014dtu} dataset.}
    \label{tab:render_quality}
\end{table}

Although our method focuses on multi-view surface reconstruction tasks, Gaussian splatting naturally supports novel view synthesis. Since we adopt a splatting process similar to 2DGS~\cite{huang20242dgs}, we compare our rendering results with 2DGS to analyze how our approach impacts rendering quality. On the DTU dataset, we use $12.5\%$ of the images for validation, while the remaining images are kept as the training set, and we report the rendering metrics including PSNR, SSIM and LPIPS.

The numerical comparisons are reported in Table \ref{tab:render_quality}. We notice a drop for PSNR on the training set but the results on test set show that our method achieves comparable rendering quality with 2DGS in novel view synthesis. The impact of our method on the rendering process primarily comes from the projection loss $L_{proj}$ that reduces the noises in the Gaussian point clouds. To further analyze the reason, we visualize the rendered images in Figure~\ref{tab:render_quality}. The scenes in the DTU dataset feature complex illumination conditions, and 2DGS tends to overfit these illumination variations, resulting in the generation of additional noisy Gaussians during the densification process. Our method smooths out these noises, which compromises the representation of the illumination. However, this smoothing removes the floating artifacts and leads to an improvement in rendering quality. Overall, our method achieves results that are quantitatively comparable to 2DGS.

\begin{figure}
  \centering
  \includegraphics[width=1.0\linewidth]{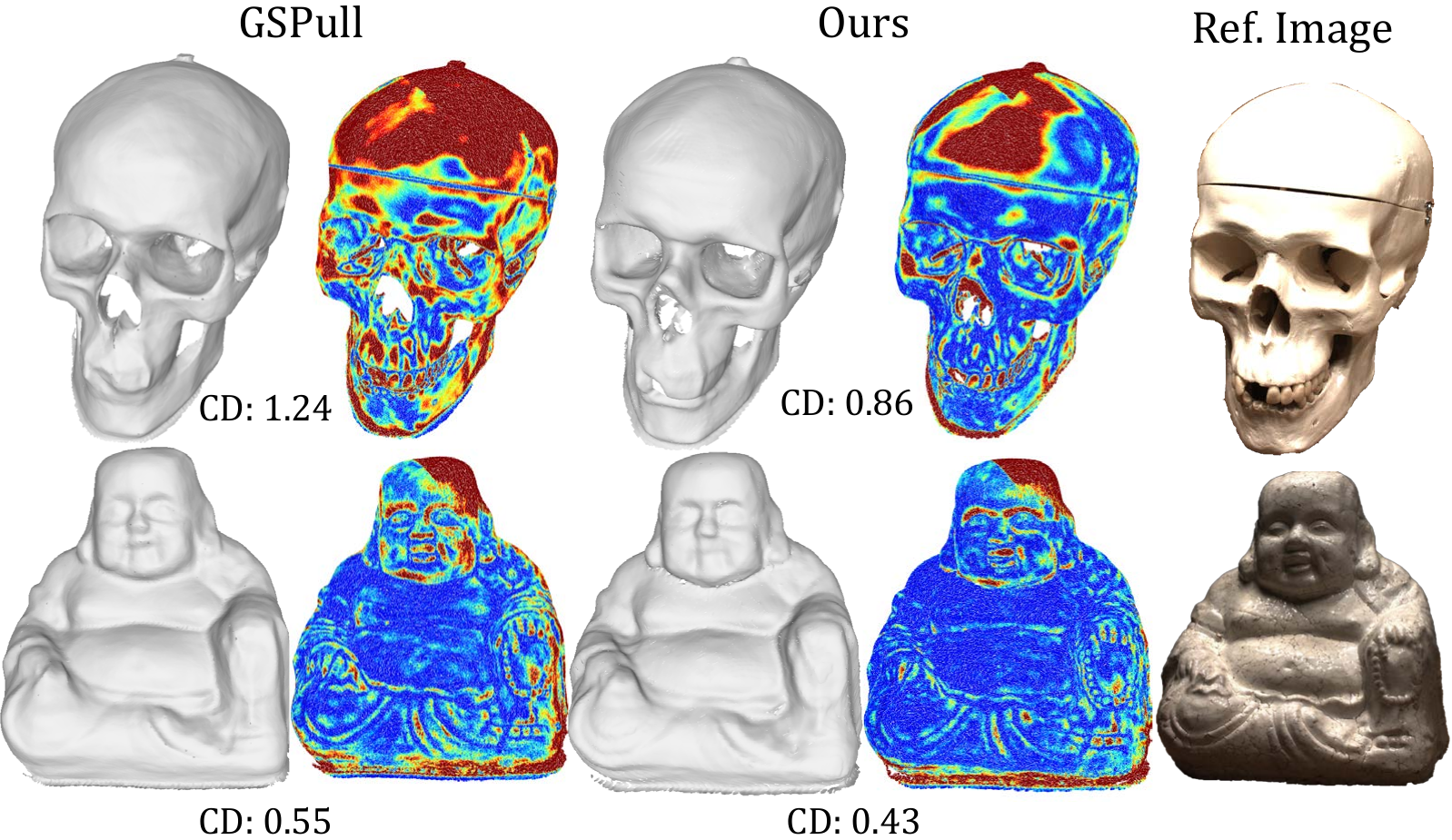}
   \caption{Visual comparisons with GSPull~\cite{zhang2024gspull} show that our method reconsturcts surfaces with less errors.}
   \label{fig:comparison_gs}
\end{figure}

\section{More Results}
We show more visual comparisons in our video. The comparisons show that our method recovers more accurate and thinner surfaces for open structures either in synthetic or real scenes. 
We visualize more reconstructed meshes on DF3D~\cite{zhu2020df3d} in Figure~\ref{fig:df3d_more} and on DTU~\cite{jensen2014dtu} in Figure~\ref{fig:dtu_more}.And we provide visual comparisons with GSPull~\cite{zhang2024gspull} on the DTU dataset shown in Fig.~\ref{fig:comparison_gs}. The results show our method produces more accurate and complete surfaces.

The detailed results of the comparison experiments and ablation studies on the DTU dataset are reported in Table \ref{tab:dtu_results_detail} and Table \ref{tab:ablation_results_detail}, respectively. 
Our method achieves a lower average Chamfer distance, indicating that the reconstructed surfaces have smaller errors and that our approach is more robust compared to baseline methods.

\section{Video}
We provide a video containing training process visualization,  field visualization, visual comparisons on all datasets, and application of point cloud deformation as a part of our supplementary materials.

{
    \small
    \bibliographystyle{ieeenat_fullname}
    \bibliography{main}
}

%% file: sec/related_work.tex
\section{Related Work}
\label{sec:related_work}

\subsection{Neural Implicit Representation}
Neural implicit representations~\cite{mescheder2019occupancy, park2019deepsdf, On-SurfacePriors, PredictiveContextPriors, noda2025bijective, zhang2025nerfprior, zhang2025monoinstance, BaoruiNoise2NoiseMapping} have shown great advantages in representing shapes using continuous functions due to their ability to represent surfaces with flexible topology in high resolutions. Typically, neural implicit functions map spatial query coordinates to occupancy probabilities \cite{mescheder2019occupancy} or signed/unsigned distances \cite{park2019deepsdf, chibane2020neural}. Neural implicit functions can be learned from various 2D or 3D surface signals, such as RGB images \cite{wang2021neus, long2023neuraludf, zhang2024vrprior}, point clouds \cite{ma2021npull, zhou2024cap, chen2024inferring, takeshi2024multipull}, binary classification labels \cite{mescheder2019occupancy}, and distance labels \cite{park2019deepsdf, chibane2020neural, li2022learning}. Among them, Neural-Pull \cite{ma2021npull} aims to pull the query points on the zero level set of the neural implicit function and achieves the learning of Signed Distance Functions (SDFs) from point clouds. The SDFs partition surfaces into exterior and interior regions, which limits the performance of such methods in modeling open surfaces. To extend the capability of implicit functions to reconstruct open surfaces, recent methods \cite{chibane2020neural, ye2022gifs, long2023neuraludf, zhou2024cap, zhou2023levelset} were proposed to predict the unsigned distances from any query points to reconstruct high quality single layer surfaces. Some methods~\cite{guillard2022meshudf, zhang2023surface, zhou2022learning, chen2022neural} extend marching cubes \cite{kobbelt2001feature, lorensen1998marching} or dual contouring \cite{ju2002dual} to efficiently and accurately extract meshes from UDFs.

\subsection{Novel View Synthesis}
Neural Radiance Fields (NeRF) \cite{mildenhall2021nerf}  have achieved promising results in novel view synthesis. The method adapts implicit field functions to encode view-dependent appearance. Specifically, NeRF maps the spatial points sampled on the ray to densities and colors with several Multi-Layer Perceptrons (MLPs), and then integrates the samples into pixel colors through volumetric rendering. Advancements~\cite{barron2021mip, barron2022mip, muller2022instant, yu2021plenoctrees, yu2021pixelnerf, jain2021putting, zhang2023fast} following the development of NeRF have further extended its capabilities.

Recently, 3D Gaussian Splatting (3DGS) \cite{kerbl20233dgs} has become an important breakthrough in the field. 3DGS represents the scene with 3D Gaussians including means, covariances, opacities and spherical harmonics parameters. The explicit representation avoids unnecessary computation cost in the empty space and achieves high quality and real-time novel view synthesis. Recent methods \cite{wu2024recent,yu2024mip-splatting, lu2024scaffold, wu20244d, tangdreamgaussian, han2024binocular, Ma2025See3D, zhou2024diffgs} extend this technique to a wide variety of fields.

\subsection{Learning Neural SDFs with Multi-view Images}
 Combining implicit representations with neural rendering, NeRF-based methods \cite{wang2021neus, oechsle2021unisurf, fu2022geo, wang2022hf} can reconstruct watertight meshes well from
 multi-view images. These methods transform occupancy values~\cite{oechsle2021unisurf} or signed distances~\cite{wang2021neus, fu2022geo, li2023neuralangelo, darmon2022improving, yariv2021volume} to density in volumetric rendering. 

Recently, attempts~\cite{Yu2024GOF, guedon2024sugar, huang20242dgs,  dai2024high} have been made to reconstruct meshes from mutiple views with 3DGS. Several methods~\cite{guedon2024sugar, dai2024high, huang20242dgs} have been developed to make 3D Gaussians approximate surfels and align with surfaces. And some methods \cite{chen2023neusg, lyu20243dgsr, yu2024gsdf, zhang2024gspull} optimize SDFs together with the 3D Gaussians. GOF~\cite{Yu2024GOF} establishes a Gaussian opacity field from 3D Gaussians and extracts the surface from the levelset. However, these methods learn SDFs to model surfaces and are limited to reconstructing watertight meshes. In contrast, we aim to handle thin and open surfaces with 3DGS, which can efficiently reconstruct non-watertight meshes. The recent method GSPull~\cite{zhang2024gspull} also pulls the queries to the zero level set to learn SDF. However, the projection can not provide enough supervision to learn correct UDF due to the complexity of gradients on the iso-surface. Therefore, we introduce self-supervision and other losses to overcome this challenge and reconstruct accurate and complete open surfaces.

\subsection{Learning Neural UDFs with Volume Rendering}
Unlike SDF modeling the surfaces as exterior and interior, UDF \cite{chibane2020neural, zhou2024cap} can handle arbitrary topologies. Recent methods \cite{long2023neuraludf, liu2023neudf, meng2023neat, deng20242sudf, zhang2024vrprior} usually learn a UDF from multi-view images with volume rendering. NeuralUDF~\cite{long2023neuraludf} flips the normal orientation behind the surface points. NeUDF~\cite{liu2023neudf} introduces a new probability density function. NeAT~\cite{meng2023neat} learns additional validity to reconstruct open surfaces from SDF. 2S-UDF~\cite{deng20242sudf} proposes a two-stage method to decouple density and weight. However, these methods need finding intersections and ray tracing in volume rendering, which leads to inefficiency. Our method is built on the point-based rendering of 3D Gaussian Splatting \cite{kerbl20233dgs} without requiring any ray tracing process, resulting in improved efficiency.

%% file: sec/experiment.tex
\section{Experiments}
\label{sec:exp}

\subsection{Experiment Settings}

\noindent\textbf{Details. }
The weights are set as $\lambda_1=0.2, \lambda_2=1.0, \lambda_3=1.0, \lambda_4=0.15$ on DTU~\cite{jensen2014dtu}, $\lambda_4=0.1$ on~DF3D~\cite{zhu2020df3d}, $\lambda_4=0.0001$ on real scans, $\lambda_5 = 1000$ on DTU, $\lambda_5 = 0$ on other scenes, and $\lambda_6 = 0.05$. We optimize the model for 30k iterations for all datasets. We use $L_{far}$ from iteration 9k to 12k, and then add $L_{near}$ and $L_{proj}$. For the self-supervision, we sample 500 Gaussian planes per batch and sample 10 root points per plane. The offset $t_b$ is sampled from a uniform distribution that is bounded by zero and $T$, and we set $T=0.01$ in DF3D dataset and $T=0.02$ in DTU dataset. Similar to NeuralUDF \cite{long2023neuraludf} and VRPrior \cite{zhang2024vrprior}, we tune the reconstruction using an additional warp loss \cite{fu2022geo, darmon2022improving} on DTU dataset. The UDF $f$ is parameterized by a 8-layer MLP with 256 hidden units and ReLU activation functions, and the activation of the last layer is an absolute value function. We apply positional encoding \cite{mildenhall2021nerf} to the input query point coordinates. We use an initial learning $1 \times 10 ^{-3}$ with cosine learning rate decay strategy for training the UDF network. We conduct all experiments on a single NVIDIA 3090 GPU.
\begin{table}
  \centering

  \begin{tabular}{@{}l|ccccc@{}}
    \toprule
    Method & 2DGS&GOF & GSPull & VRPrior & Ours \\
    \midrule
    Average& 0.80 & 0.74& 0.75&0.71 &\textbf{0.68}\\
    \bottomrule
  \end{tabular}
  \caption{Numerical comparisons with 2DGS~\cite{huang20242dgs}, GOF~\cite{Yu2024GOF}, GSPull~\cite{zhang2024gspull} and VRPrior~\cite{zhang2024vrprior} in terms of CD on DTU~\cite{jensen2014dtu} dataset. Detailed comparisons can be found in the appendix.}
  \label{tab:dtu_results}
  \vspace{-15pt}
\end{table}

\noindent\textbf{Datasets and Evaluation Metrics.}  We evaluate the proposed method on DeepFashion3D (DF3D) \cite{zhu2020df3d} dataset,  DTU \cite{jensen2014dtu} dataset, NeUDF \cite{liu2023neudf} dataset, and our real-captured dataset. For DF3D dataset, we use the same 12 garments as previous methods \cite{long2023neuraludf, zhang2024vrprior}, each garment is scanned with 72 images in a resolution of $1024 \times 1024$ and is provided with the ground truth point cloud for evaluation. For DTU \cite{jensen2014dtu} dataset, we use the widely used 15 scenes that are all watertight and each scene contains 49 or 64 images in a resolution of $1600 \times 1200$. We use two real scans in NeUDF \cite{liu2023neudf} dataset, and captures four real scenes. In our experiments, we train our models without mask supervision in all datasets. For a fair comparison, we use the MeshUDF \cite{guillard2022meshudf} algorithm to extract open surfaces from unsigned distance fields like previous methods \cite{long2023neuraludf, zhang2024vrprior, deng20242sudf, liu2023neudf}, and use the Chamfer Distance (CD) as the metric for DF3D dataset and DTU dataset that provide ground truth. 

\noindent\textbf{Baselines.} We compare the proposed method with the following state-of-the-art methods: 1) SDF-based surface reconstruction methods including NeuS~\cite{wang2021neus}, 2DGS \cite{huang20242dgs}, and GOF \cite{Yu2024GOF}, and 2)UDF-based surface reconstruction methods for open surfaces including NeuralUDF \cite{long2023neuraludf}, 2S-UDF~\cite{deng20242sudf}, and VRPrior \cite{zhang2024vrprior}. For the open surface dataset DF3D~\cite{zhu2020df3d}, 
we trained GOF~\cite{Yu2024GOF} and 2S-UDF~\cite{deng20242sudf} with the default parameters. Since we share Gaussian optimization parameters with 2DGS~\cite{huang20242dgs}, we keep these parameters the same. The other quantitative metrics are borrowed from the original papers.

\subsection{Evaluations}

\textbf{Comparisons in Reconstructing Open Surfaces.} 
We evaluate our method on the DF3D\cite{zhu2020df3d} dataset which includes shapes with open surfaces. The CD $(\times 10^{-3})$ in Table \ref{tab:df3d_results} indicates that we achieve the best performance compared to baseline methods. The reconstruction errors with SDF-based baselines including NeuS~\cite{wang2021neus}, 2DGS~\cite{huang20242dgs}, and GOF~\cite{Yu2024GOF} are large because they try to either wrap the surface with closed mesh or excessively smooth out the details on the clothing.
The visual comparisons in Figure \ref{fig:df3d_vis} show that our method can reconstruct open surfaces with more details. The methods 2DGS and GOF inherit the shortcoming of SDF-based methods 
which learn to reconstruct closed surfaces. This results double-layered faces and increases the reconstruction errors. The UDF-based baselines reconstruct the open surface correctly, but they fail to capture details, resulting in over-smoothed results. Thanks to the quick convergence of 3D Gaussian splatting, the speed of training our method can be much faster than the NeRF-based methods for open surface reconstruction.

\noindent\textbf{Comparisons in Reconstructing Closed Surfaces.}
We further conduct evaluations on the DTU dataset, and report the quantitative and visual comparisons in Table \ref{tab:dtu_results} and Figure \ref{fig:dtu_vis}, respectively. Our method achieves the best performance in terms of average CD compared among baseline methods, demonstrating its overall robustness. 
The complex gradients near the surface make the learning of UDF more challenging than SDF. Without assuming closed surfaces, our method still achieves comparable results or even better results in some scenes to SDF-based methods that are specifically designed for closed surfaces. Moreover, our approach achieves better quantification on some relatively complex shapes than baseline methods. As shown in the error map in Figure~\ref{fig:dtu_vis}, our method accurately reconstructs surface even with complex light conditions. The underlying reason is that the geometric information of the UDF is derived from the positions of Gaussians, making it less sensitive to appearance attributes like opacity.

\begin{figure}
  \centering
  \includegraphics[width=1.0\linewidth]{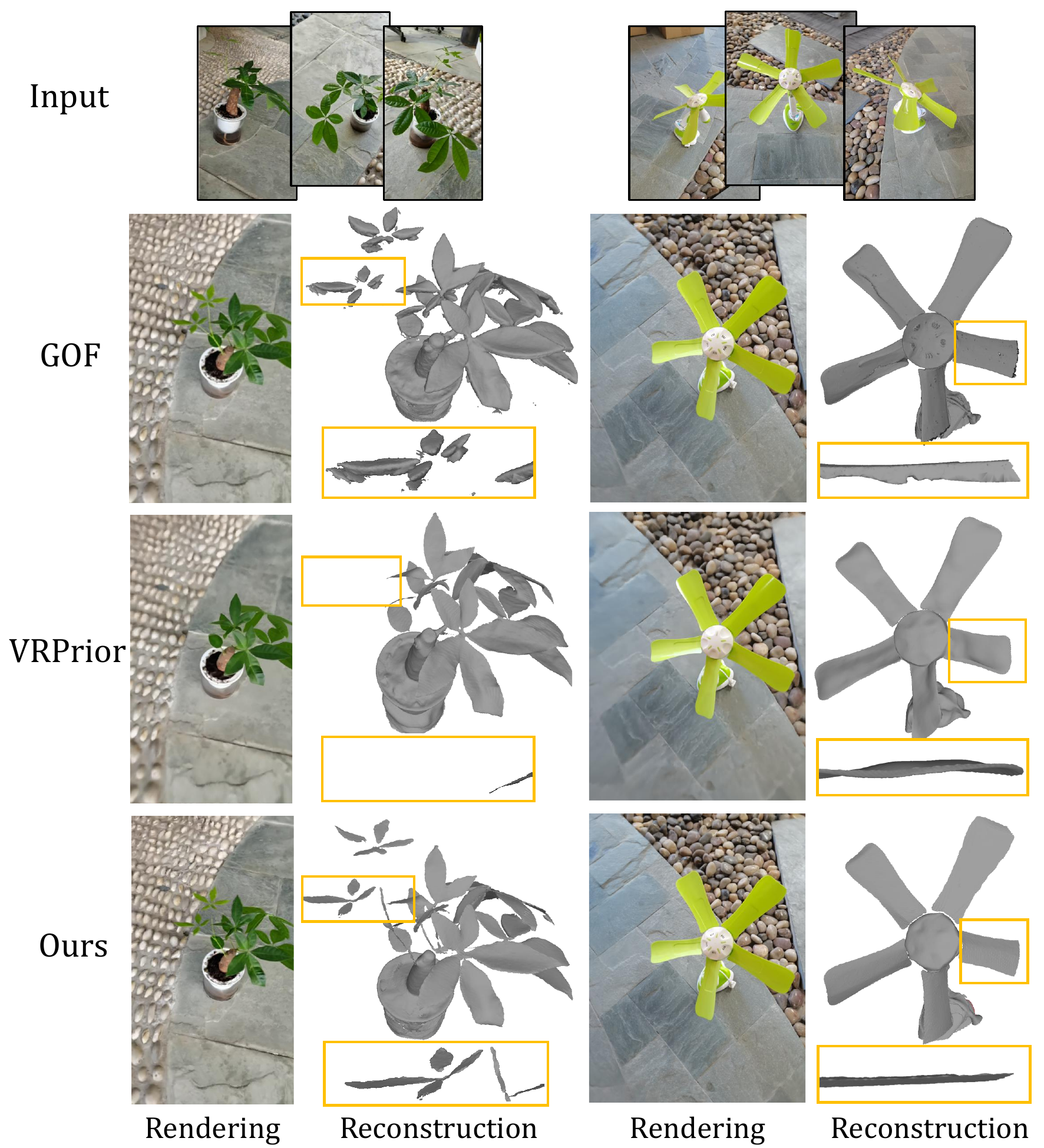}

   \caption{The reconstruction results on NeUDF \cite{liu2023neudf} dataset. Our method accurately reconstructs the open surfaces in real scans.}
   \label{fig:neudf_vis}
\end{figure}

\begin{figure}
  \centering
  \includegraphics[width=1.0\linewidth]{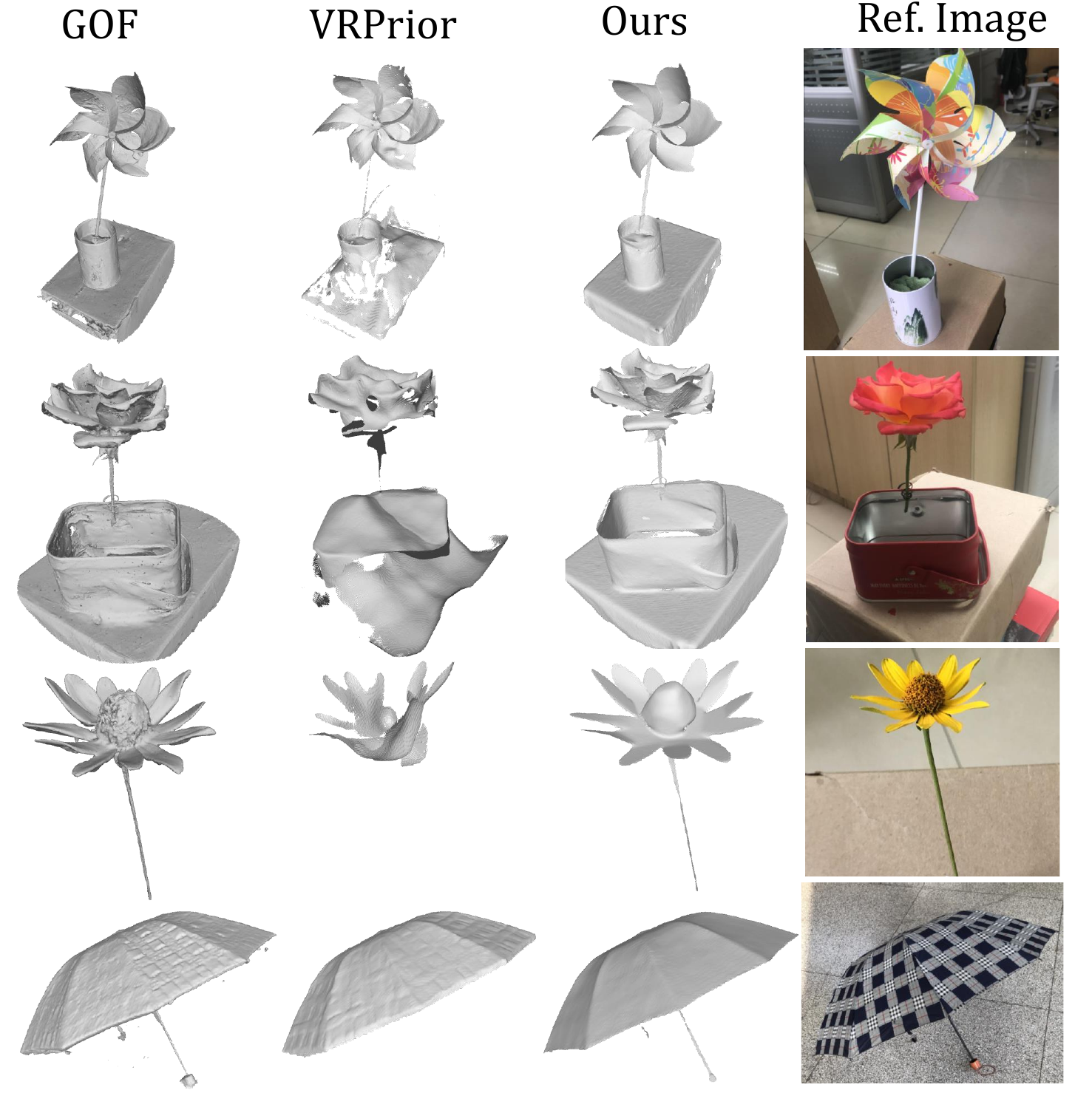}

   \caption{The reconstruction results on real scans. Our method reconstructs more accurate and complete surfaces.}
   \vspace{-15pt}
   \label{fig:real_vis}
\end{figure}

\begin{table}
  \centering
  \begin{tabular}{@{}l|cccc|c@{}}
    \toprule
    Settings & Far & Near & Proj & Warp & CD $\downarrow$ \\
    
    \midrule
    Only Far& \checkmark& & & &0.99\\
    Far$\And$Near& \checkmark& \checkmark &&&0.78\\
    Far$\And$Proj&\checkmark& & \checkmark &&0.88\\
    w/o Warp&\checkmark&\checkmark&\checkmark&&0.74\\
    w/o Near& \checkmark& & \checkmark & \checkmark&0.77 \\
    w/o Proj&\checkmark&\checkmark&&\checkmark&0.76\\
    Full Model&\checkmark&\checkmark&\checkmark&\checkmark&0.68\\
    \bottomrule
  \end{tabular}
  \caption{Ablation studies on DTU dataset. The results show that all designs in our method are effective.}
  \label{tab:ablation_study}
\end{table}

\begin{figure}
  \centering
  \includegraphics[width=1.0\linewidth]{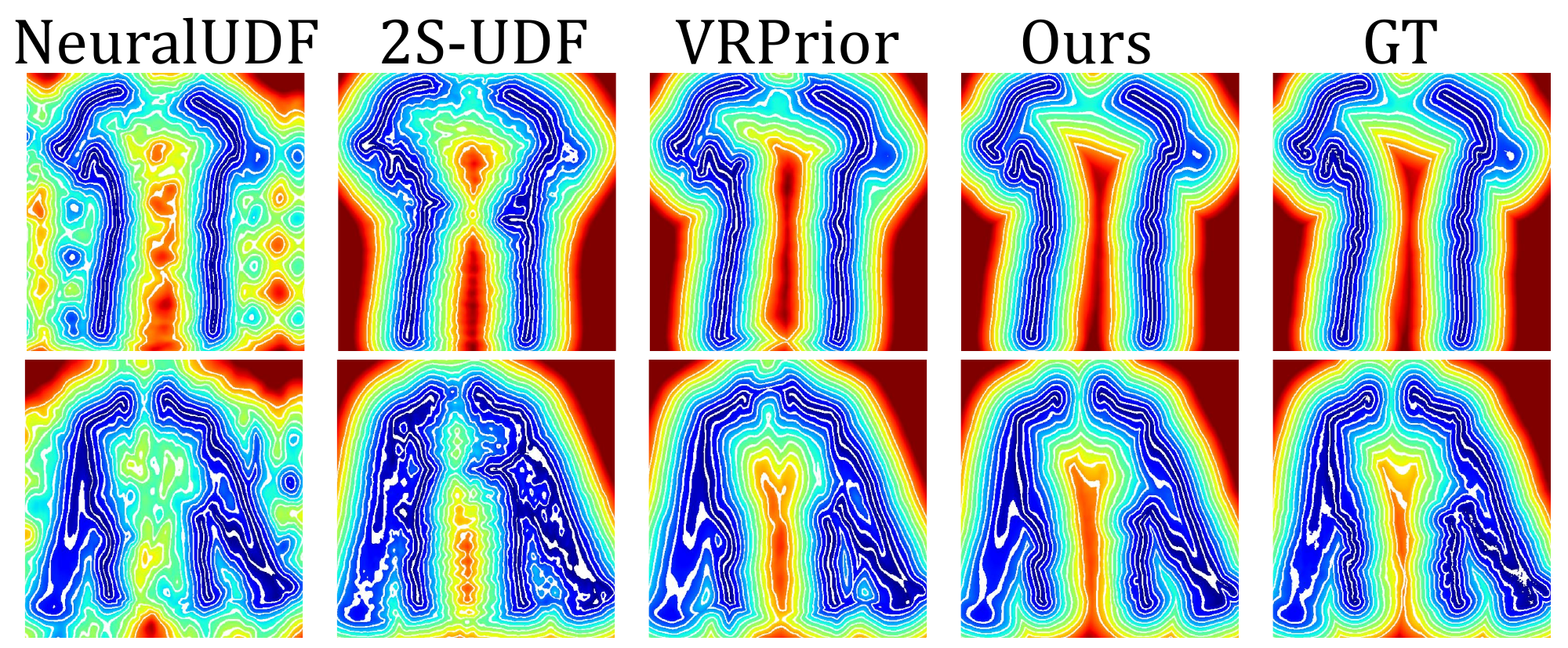}
   \caption{The UDFs learned with different methods. Our method learned more complete and smoother level sets in the field.
   }
   \vspace{-5pt}
   \label{fig:field_vis}
\end{figure}

\begin{figure*}
  \centering
  \includegraphics[width=1.0\linewidth]{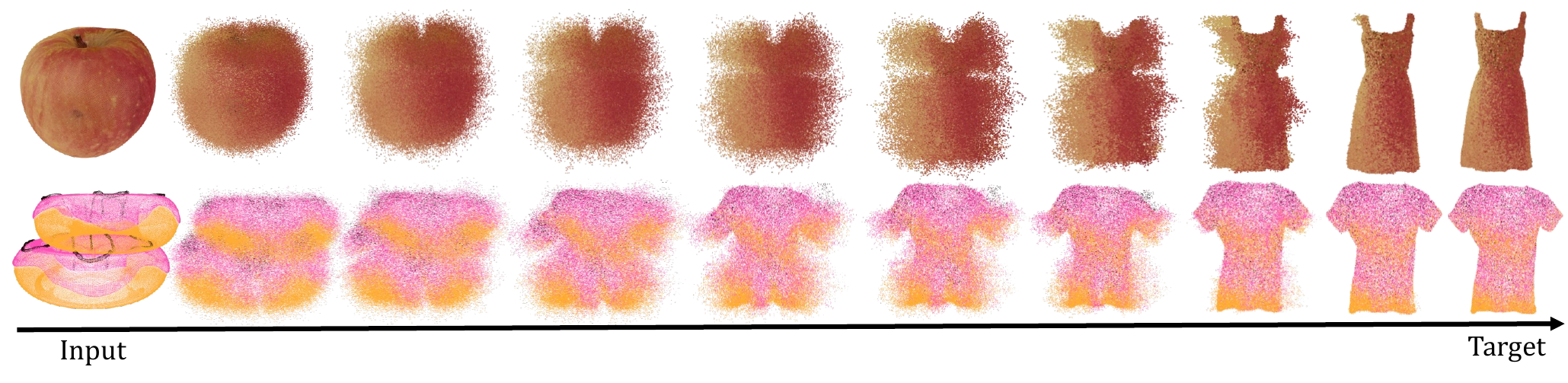}
   \caption{Point cloud deformation in the learned unsigned distance field. The accurate field can deform point clouds with any shapes (such as apple and donut) into the target shapes represented by the UDFs.}
   \vspace{-0.2cm}
   \label{fig:deform}
\end{figure*}

\noindent\textbf{Results on Real Scans.}
We first conduct evaluation on the public real-captured NeUDF \cite{liu2023neudf} dataset. As shown in Figure \ref{fig:neudf_vis}, our method can reconstruct extremely flat and thin surfaces. Due to the detail-capturing capability of Gaussian Splatting, our method achieves a more complete geometry reconstruction compared to the NeRF-based state-of-the-art VRPrior \cite{zhang2024vrprior}, such as the plant leaves, even if it uses additional data-driven learned priors. We further report our results on our self-captured four scenes with thin and open surfaces. As shown in Figure \ref{fig:real_vis}, VRPrior~\cite{zhang2024vrprior} struggles to reconstruct correct structures for objects with relatively simple textures, and GOF~\cite{Yu2024GOF} reconstructs double-layer surfaces without smoothness. Instead, our method can reconstruct more complete, accurate, and smoother meshes.

\begin{figure}
  \centering
  \includegraphics[width=1.0\linewidth]{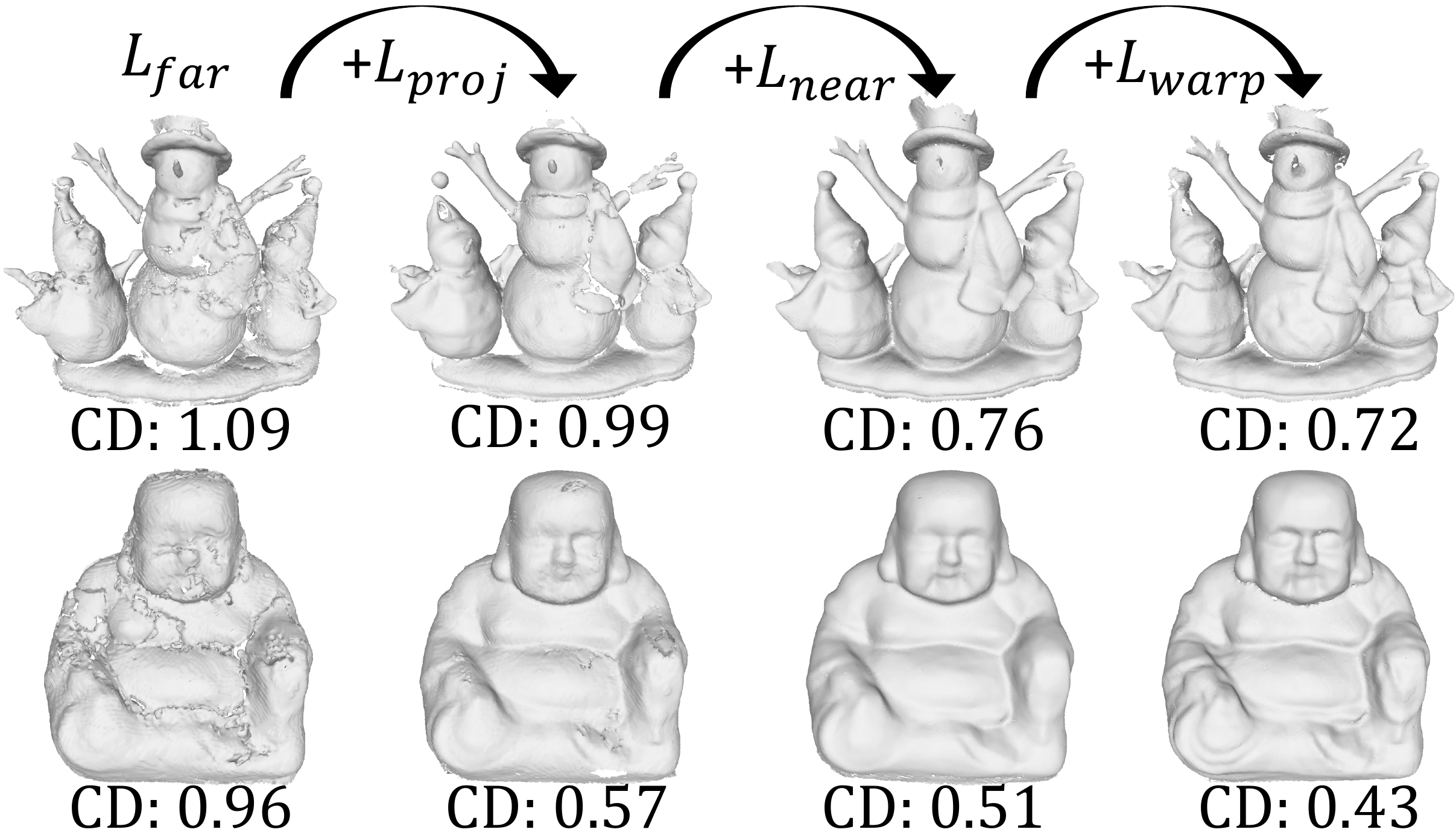}
   \caption{Visual comparisons with each of our constraints. The results show all components in our method are critical for our accurate surface reconstruction.}
   \vspace{-10pt}
   \label{fig:loss_process}
\end{figure}

\subsection{Visual Analysis in Unsigned Distance Fields}

\noindent\textbf{Visualization of Unsigned Distance Fields. }We visualize the learned unsigned distance fields in Figure~\ref{fig:field_vis}. 
We use the unsigned distances from UDFs learned by different methods and map these distances in colors. Points near the surface are close to blue, while points far from the surface are close to red. NeuralUDF~\cite{long2023neuraludf} learns zero UDF values far from the surface, which increases the difficulty of convergence. 2S-UDF~\cite{deng20242sudf} learns a complex function close to the surface due to overfitting on textures. With the help of the depth prior, VRPrior~\cite{zhang2024vrprior} learns better fields. However, it fails to capture the correct boundaries and almost closes the adjacent open surfaces. Our method learns the most accurate implicit functions without any extra prior.

\noindent\textbf{Point Cloud Deformation. }With the learned unsigned distance function, we can obtain the distance and the direction pointing to the surface for any point. Therefore, the UDF %is also a deformation function that 
can deform source point clouds into the shape represented by the UDF. As shown in Figure \ref{fig:deform}, we gradually pull the input point clouds into the garments with Eq.~(\ref{Eq:pull}), which validates that the implicit function has learned correct surface information at any point in space.

\subsection{Ablation Studies}
We conduct ablation studies on the DTU
 dataset \cite{jensen2014dtu} to show the impact of each module on the performance, and the full quantitative results are reported in Table \ref{tab:ablation_study}.

Firstly, we try to learn the unsigned distance fields directly from the Gaussian point clouds, which is similar to the target of point cloud reconstruction \cite{ma2021npull, zhou2024cap}. As shown by the row ``Only Far'' in Table \ref{tab:ablation_study}, the performance drops significantly and the reason is that the point clouds of Gaussians are noisy, sparse and uneven, which cannot provide accurate geometry information. Overfitting a low-quality point cloud results in a poor surface, as shown in the first picture in Figure \ref{fig:loss_process}. We also combine the $L_{far}$ with $L_{proj}$ and $L_{near}$ respectively. The results in ``Far$\And$Near'' and ``Far$\And$Proj'' show that both losses are critical for the accurate reconstruction and $L_{near}$ plays a more important role.

To show how each loss affects our method, we remove the terms one by one and report the metrics as ``w/o Near'', ``w/o Proj'' and ``w/o Warp''. The results show that each loss plays a positive role in the final result, verifying the effectiveness of different parts of our method. Besides, removing the $L_{near}$ loss leads to the largest drop in average metrics, which also proves that the self-supervision loss provides the most important information for learning UDFs. 

We gradually add different losses in the order of $L_{far}$, $L_{proj}$, $L_{near}$, and $L_{warp}$, and show the changes in results in Figure \ref{fig:loss_process}. Projecting Gaussians to the surface helps to learn a smooth surface, and self-supervision can fill the holes in the meshes. The warp loss captures more details. All loss terms contribute to more accurate surface reconstruction.

\textbf{Limitations. }Compared to SDF-based reconstruction methods, our approach demonstrates reduced performance in reconstructing textureless structures. This limitation arises from the high flexibility of UDF, which introduces complexities into the optimization process. Moreover, extracting surfaces from UDF fields is still an ongoing challenge~\cite{zhou2024cap, zhang2023surface}, which constrains the quality of the reconstructed open mesh. These factors result in a lack of detail in the surfaces reconstructed by our method, particularly for complex structures. In future work, incorporating additional priors, such as normals, masks, and depth, could help capture higher-frequency signals. Furthermore, integrating our approach with the latest UDF extraction methods~\cite{zhang2023surface, chen2022neural} may also enhance the quality of the reconstructed mesh.